\documentclass[11pt]{article}

\usepackage[final]{acl}

\usepackage{times}
\usepackage{latexsym}
\usepackage{amsmath, amsfonts, amssymb} 
\usepackage{cleveref}
\usepackage{booktabs}   
\usepackage{multirow}
\usepackage{array}    
\usepackage{siunitx}
\usepackage{algorithm}
\usepackage{algpseudocode}
\usepackage{xcolor}
\usepackage{listings}
\usepackage{bookmark}
\usepackage{todonotes}
\usepackage{pifont}
\usepackage{bbding}

\lstdefinestyle{eacl}{
  basicstyle=\ttfamily\small,
  breaklines=true,
  frame=single,
  rulecolor=\color{black!30},
  backgroundcolor=\color{gray!5},
  columns=fullflexible,
  keepspaces=true,
  captionpos=b,
  abovecaptionskip=1ex,
  belowskip=2ex,
  showstringspaces=false
}

\usepackage[T1]{fontenc}

\usepackage[utf8]{inputenc}

\usepackage{microtype}

\usepackage{inconsolata}

\usepackage{graphicx}

\usepackage{xcolor}
\usepackage{inconsolata} 
\usepackage{tcolorbox}
\tcbuselibrary{skins}

\definecolor{promptbg}{HTML}{F8F8F8}
\definecolor{promptframe}{HTML}{DCDCDC}
\definecolor{prompttitle}{HTML}{444444}

\newtcolorbox{promptbox}[1]{
  colback=promptbg,
  colframe=promptframe,
  fonttitle=\bfseries\color{prompttitle},
  left=6mm,
  right=6mm,
  top=4mm,
  bottom=4mm,
  arc=2mm,
  title=#1,
  fontfamily=\ttfamily, 
  fontsize=\small,
  boxrule=0.5pt,
  sharp corners
}
\title{AutoBench: Automating LLM Evaluation through Reciprocal Peer Assessment}

\author{
  \textbf{Dario Loi\textsuperscript{1}},
  \textbf{Elena Maria Muià\textsuperscript{2}},
  \textbf{Federico Siciliano\textsuperscript{2}},
  \textbf{Giovanni Trappolini\textsuperscript{2}},
\\
  \textbf{Vincenzo Crisà\textsuperscript{2}},
  \textbf{Peter Kruger\textsuperscript{3}},
  \textbf{Fabrizio Silvestri\textsuperscript{2}},
\\
\\
  \textsuperscript{1}Department of Computer Science, Sapienza University of Rome\\
  \textsuperscript{2}Department of Computer, Control, and Management Engineering, Sapienza University of Rome\\
  \textsuperscript{3}eZecute S.R.L.
\\
  \small{
    \textbf{Correspondence:} \href{mailto:loi.1940849@studenti.uniroma1.it}{loi.1940849@studenti.uniroma1.it}
  }
}

\begin{document}
\maketitle
\begin{abstract}
We present AutoBench, a fully automated and self-sustaining framework for evaluating Large Language Models (LLMs) through reciprocal peer assessment. This paper provides a rigorous scientific validation of the AutoBench methodology, originally developed as an open-source project by eZecute S.R.L.. Unlike static benchmarks that suffer from test-set contamination and limited adaptability, AutoBench dynamically generates novel evaluation tasks while models alternately serve as question generators, contestants, and judges across diverse domains. An iterative weighting mechanism amplifies the influence of consistently reliable evaluators, aggregating peer judgments into consensus-based rankings that reflect collective model agreement. Our experiments demonstrate strong correlations with established benchmarks including MMLU-Pro and GPQA (respectively 78\% and 63\%), validating this peer-driven evaluation paradigm. The multi-judge design significantly outperforms single-judge baselines, confirming that distributed evaluation produces more robust and human-consistent assessments. AutoBench offers a scalable, contamination-resistant alternative to static benchmarks for the continuous evaluation of evolving language models.
\end{abstract}

\section{Introduction}

Evaluating Large Language Models (LLMs) remains a central challenge in natural language processing. While traditional benchmarks such as MMLU \cite{hendrycks2021measuringmassivemultitasklanguage} and MATH \cite{math_dataset} have been instrumental in measuring performance across a range of subjects, they are inherently static: the task sets are fixed, models can overfit to them, and their diagnostic power diminishes as LLMs rapidly improve. 

This paper provides a rigorous scientific validation of the AutoBench framework, an open-source, self-sustaining benchmark for LLMs originally developed by eZecute S.R.L. and detailed publicly\footnote{See the official project website \cite{autobench2024project}, the Hugging Face organization \cite{autobench2024hf}, and the initial release announcement \cite{kruger2025autobench}}. This public-facing project has already demonstrated the methodology's effectiveness at scale; its third public run, for instance, evaluated 33 models over 400 iterations, collecting $\sim$300,000 individual ranks and showing exceptionally high correlation with established benchmarks, such as 92.17\% with Artificial Analysis Intelligence Index, and 86.85\% with LMSYS Chatbot Arena \cite{kruger2025autobenchrun3}. This paper's contribution is to formalize and analyze the system's core methodology, built on a fully automated framework of reciprocal peer evaluation. In this system, models autonomously conduct the entire evaluation lifecycle, from generating tasks to delivering final judgments. The process begins with the collaborative creation of diverse tasks across varying topics and difficulty levels. Each model then answers all tasks, producing a shared pool of responses for assessment. In the core peer-evaluation phase, models judge both the tasks and answers of their peers, forming a directed, weighted graph where nodes represent models and edges encode assigned scores. An iterative weighting algorithm aggregates these judgments into authority scores, amplifying the influence of consistent evaluators and establishing a dynamic, self-regulating standard for model assessment.
    
The benchmark is \textit{dynamic}, generating new prompts each iteration; \textit{reciprocal}, as models act as both examiners and reviewers; and \textit{consensus-driven}, with scores reflecting weighted collective agreement. This automation eliminates human supervision, prevents test contamination, and yields stable, interpretable leaderboards beyond the limits of fixed datasets or single-judge setups.

Our contributions are threefold:
\begin{enumerate}
    \item A peer-evaluation protocol where LLMs both generate and assess questions and answers across diverse topics and difficulty levels.  
    \item An iterative weighting algorithm that aggregates peer judgments into robust, consensus-based rankings.  
    \item Empirical validation showing coherent, adaptive leaderboards, establishing a foundation for continuous, self-sustaining LLM evaluation.  
\end{enumerate}

\section{Related Works}
The landscape of LLM evaluation has evolved from static, human-curated benchmarks to more dynamic and model-driven approaches. We categorize existing work into three main areas, highlighting the progression toward the self-evolving, reciprocal peer evaluation framework we propose.

Traditional LLM evaluation has largely relied on static, ground-truth benchmarks designed to measure performance across diverse domains. These include well-established datasets like MMLU~\cite{hendrycks2021measuringmassivemultitasklanguage}, a comprehensive 57-task multiple-choice suite, and MATH~\cite{math_dataset}, comprising $12,500$ competition-level problems. Further examples encompass specialized benchmarks such as GSM-8K~\cite{cobbe2021gsm8k} for mathematical reasoning, HumanEval~\cite{li2024humaneval} for code generation, and broader assessments GPQA~\cite{rein2024gpqa}. While instrumental for quantifying factual knowledge and problem-solving abilities, these static benchmarks face inherent limitations: they risk test set contamination as models are increasingly trained on web corpora, and their diagnostic power diminishes as LLMs improve, often struggling to provide fine-grained insights into emergent capabilities or open-ended generation.

To overcome the scalability and contamination issues of static benchmarks, a significant body of recent work leverages LLMs themselves as evaluators. This paradigm broadly divides into single-judge and multi-agent approaches. Single-judge methods typically employ a strong, often proprietary, LLM (e.g., GPT-4~\cite{openai2023gpt4}) to assess outputs. For instance, MT-Bench~\cite{zheng2023judgingllmasajudgemtbenchchatbot} utilizes GPT-4 to compare answers to multi-turn queries, demonstrating substantial agreement with human preferences. Other efforts train LLMs for prompt-specific rankings (Prompt-to-Leaderboard\cite{frick2025prompttoleaderboard}) or integrate human feedback into large-scale hybrid evaluations like Chatbot Arena~\cite{chiang2024chatbot}. Building on this, multi-agent schemes simulate panels of evaluators to achieve more robust judgments. ChatEval~\cite{chan2023chatevalbetterllmbasedevaluators} involves multiple LLM "personas" or paired reviewers, often weighting judgments from more reliable agents. Auto-Arena~\cite{zhao2024autoarenaautomatingllmevaluations} introduces a multi-stage pipeline involving an examiner, debate, and judge committee, boosting alignment with human preferences over static tests. While these methods advance automated evaluation, they often rely on a centralized ``strong'' judge or pre-defined roles, and the aggregation of judgments can be based on fixed or learned weights rather than a dynamic, network-driven authority.

A complementary direction emphasizes the necessity of dynamic and evolving benchmarks to counter contamination and keep pace with rapid LLM advancements. Initiatives like DynaBench \cite{kiela2021dynabenchrethinkingbenchmarkingnlp} and LiveBench \cite{white2025livebenchchallengingcontaminationlimitedllm} advocate for ``living'' evaluations that continuously adapt. BenchBuilder~\cite{li2024crowdsourceddatahighqualitybenchmarks} exemplifies this by filtering Chatbot Arena queries to create Arena-Hard-Auto~\cite{li2024crowdsourceddatahighqualitybenchmarks}, a 500-task benchmark achieving high agreement with human rankings at low cost. While these dynamic approaches mitigate dataset staleness, they remain limited in several ways: DynaBench continues to rely on human-in-the-loop task creation, LiveBench employs a fixed set of evaluator models, and Arena-Hard-Auto constructs its benchmark from filtered human–LLM interaction logs rather than autonomously generated tasks. 

In contrast, our framework removes these dependencies through a fully automated, reciprocal peer-evaluation process. All participating models simultaneously generate tasks, produce answers, and assess one another, with judgments aggregated via an iterative weighting mechanism that amplifies reliable evaluators. This design achieves end-to-end automation without fixed roles or human oversight, yielding a self-evolving, contamination-resistant benchmark that continuously adapts alongside the LLMs it measures.

\section{Methods}\label{sec:methods}

We introduce \textsc{AutoBench}, a dynamic and self-contained framework for evaluating a set of language models, denoted by $\mathcal{M} = \{M_1, \dots, M_n\}$. 
Each model can take on three distinct roles during evaluation: \emph{task generator}, \emph{contestant}, and \emph{judge}. 
The system evolves over a series of $T$ iterations, progressively refining model evaluations to converge towards a stable consensus ranking.

\begin{figure*}[tbh]
  \includegraphics[width=0.98\textwidth]{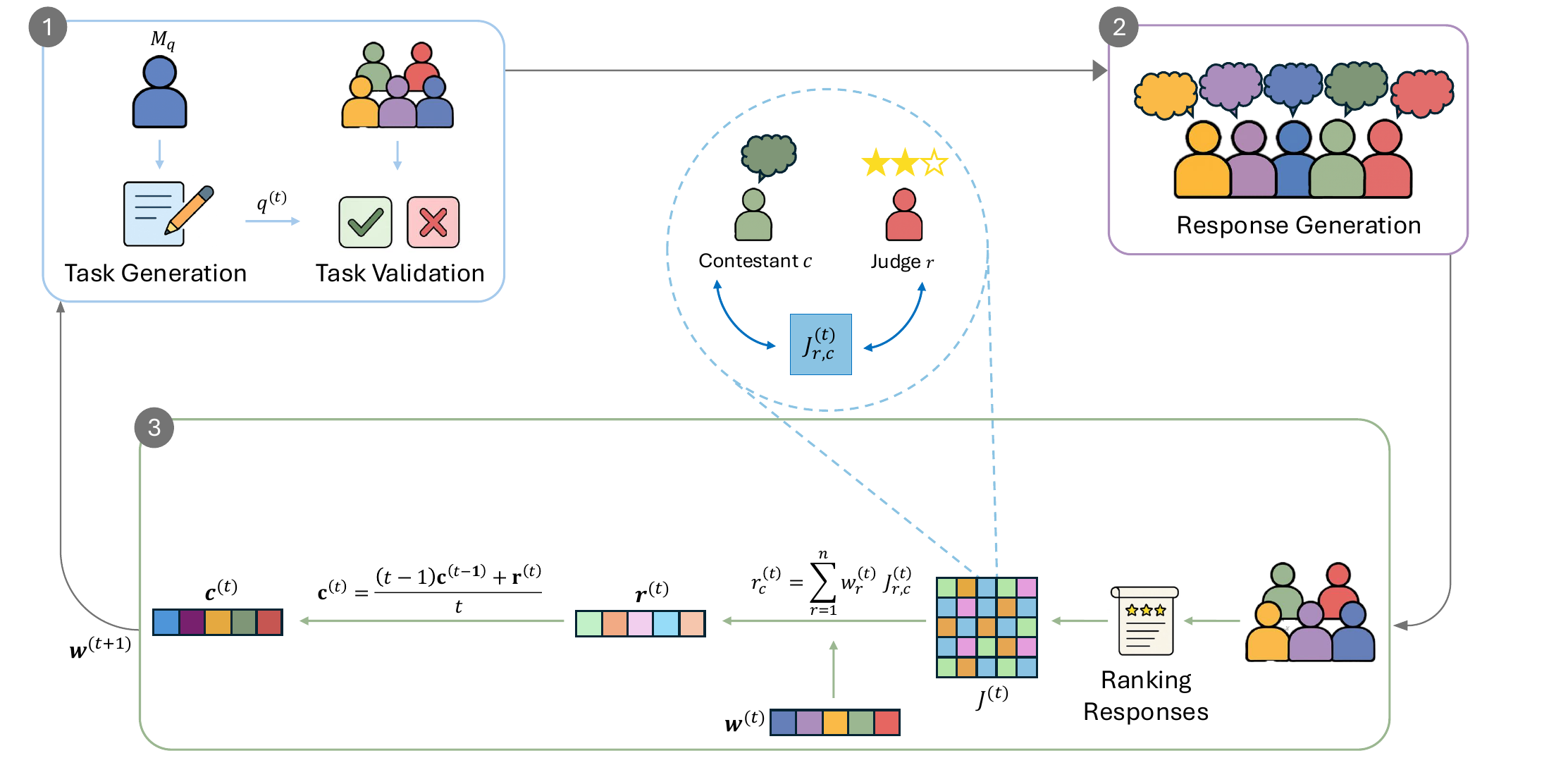}
  \caption{Diagram of the \textsc{AutoBench} framework.}
  \label{fig:cross_evaluation_summary}
\end{figure*}

At iteration $t$, the state of the system is represented by a judging weight vector $\mathbf{w}^{(t)} = (w_1^{(t)}, \dots, w_n^{(t)})$, 
which quantifies the relative authority of each model when acting as a judge. 
Weights are initialized uniformly, such that $w^{(1)}_k = \frac{1}{n}$ for all models $k \in \{1, \dots, n\}$.

At the start of each iteration, one model $M_q$ is sampled from $\mathcal{M}$ to serve as the \emph{task generator}. 
To ensure a comprehensive evaluation across diverse capabilities, tasks are drawn from a discrete set of predefined categories (see \Cref{sec:experiments}). 
The generated task, denoted by $q^{(t)}$, undergoes a quality assurance phase in which all models provide preliminary assessments. 
A task is accepted only if its mean and median weighted scores, computed using the current weight vector $\mathbf{w}^{(t)}$, exceed predefined thresholds $\lambda_{\text{mean}}$ and $\lambda_{\text{median}}$, respectively. 
If no acceptable task is produced after $k$ attempts, the process restarts with a newly sampled generator.

Once a task is accepted, every model in $\mathcal{M}$ participates in two roles: 
first as a \emph{contestant}, producing a response, and then as a \emph{judge}, evaluating the responses of all contestants. 
We represent these evaluations in a \emph{judgment matrix} $J^{(t)} \in \mathbb{R}^{n \times n}$, 
where each row corresponds to a judge and each column to a contestant. 
The entry $J^{(t)}_{r,c}$ denotes the score given by the $r$-th judge to the $c$-th contestant. 
Because all models act in both roles, $J^{(t)}$ provides a complete pairwise assessment of the system.

At the conclusion of iteration $t$, an aggregate score vector $\mathbf{r}^{(t)}$ is computed for the contestants. 
Each element of $\mathbf{r}^{(t)}$ is obtained as the weighted sum of the corresponding column of $J^{(t)}$:
\begin{equation}
    r_c^{(t)} = \sum_{r=1}^{n} w_r^{(t)} \, J^{(t)}_{r,c}.
\end{equation}
This instantaneous score vector is then used to update the models' cumulative performance scores, $\mathbf{c}^t$, which are maintained as a running average:
\begin{equation}\label{eq:cum_score}
    \mathbf{c}^{(t)} = \frac{(t-1)\mathbf{c}^{(t-1)} + \mathbf{r}^{(t)}}{t}
\end{equation}
The resulting values, in turn, normalized to derive the judging weights for the subsequent iteration:
\begin{equation}\label{eq:weight_norm}
    w_j^{(t+1)} = \frac{c_j^{(t)}}{\sum_{k=1}^n c_k^{(t)}}
\end{equation}
This iterative re-weighting mechanism ensures that models demonstrating consistently high performance gain greater influence in the evaluation process, thereby enhancing the robustness of the \textsc{AutoBench} framework. A detailed pseudo-code of the full procedure is available in \Cref{alg:autobench}.

\section{Experiments}\label{sec:experiments}

\begin{table*}[tbh]
\centering
\small
\setlength{\tabcolsep}{8pt}
\renewcommand{\arraystretch}{1.12}
\resizebox{0.95\textwidth}{!}{%
\begin{tabular}{lcccc}
\toprule
\multirow{2}{*}{\textbf{Compared benchmark}} &
  \multicolumn{2}{c}{\textbf{Multi-judge (\textsc{AutoBench})}} &
  \multicolumn{2}{c}{\textbf{Single-judge (\texttt{gpt-oss-20b})}} \\
\cmidrule(lr){2-3} \cmidrule(lr){4-5}
 & \textbf{Kendall’s $\boldsymbol{\tau}$} & \textbf{Spearman’s $\boldsymbol{\rho}$}
 & \textbf{Kendall’s $\boldsymbol{\tau}$} & \textbf{Spearman’s $\boldsymbol{\rho}$} \\
\midrule
MMLU-Pro       & $\mathbf{0.64}$ & $\mathbf{0.78}$ & $0.45 $ & $0.61 $ \\
GPQA           & $\mathbf{0.52}$ & $\mathbf{0.63}$ & $0.45 $ & $0.59 $ \\
\bottomrule
\end{tabular}%
}
\caption{\textbf{Correlation analysis: multi-judge vs single-judge configurations.}
Kendall’s $\tau$ and Spearman’s $\rho$ computed between \textsc{AutoBench} model rankings and established benchmarks. The left block corresponds to the full multi-judge pipeline; the right block corresponds to an ablation using only \texttt{gpt-oss-20b} as judge.}
\label{tab:merged_corr_ablation}
\end{table*}

\paragraph{Experimental Setup}  
We evaluate the \textsc{AutoBench} framework presented in \Cref{sec:methods} on a suite of twelve widely available open-source models, detailed in \Cref{tab:model-card}. All models were sourced from the Hugging Face Hub\footnote{\href{https://huggingface.co}{https://huggingface.co}} and are governed by open licenses that permit research use. Our methodology adheres to their intended use as instruction-following and chat models.
The framework was executed for $T=40$ iterations, with task quality enforced via score thresholds of $\lambda_{mean} = 3.5$ and $\lambda_{median} = 3.0$, allowing up to $k=3$ generation retries. For all model responses, we used a temperature of $\tau = 0.8$ and nucleus sampling with $p_{top}=0.9$ to encourage diverse yet coherent outputs~\cite{nucleus_sampling}.

To validate the robustness of \textsc{AutoBench}, we measure how closely its emergent model rankings align with established human and task-based benchmarks, including MMLU-Pro~\cite{hendrycks2021measuringmassivemultitasklanguage}, MATH~\cite{math_dataset}, and GPQA~\cite{rein2024gpqa}. 
This comparison assesses whether \textsc{AutoBench} produces rankings that are consistent with external evaluations of model capability. 
To quantify this alignment we use Kendall’s $\tau$ and Spearman’s $\rho$~\cite{Spearman2015ThePA,kendall} as correlation statistics, the result of this study are reported in~\Cref{tab:merged_corr_ablation}.

To isolate the impact of the multi-judge design, we perform an ablation in which only the best open-source model, \texttt{gpt-oss-20b}, serves as judge. 
This single-judge configuration is compared directly with the full multi-judge pipeline in \Cref{tab:merged_corr_ablation}, enabling a direct assessment of how collective judgment influences benchmark correlations.

Our task-generating prompts are conditioned to generate tasks belonging to a diverse set of predefined categories: logic, coding, technology, history, science, general culture, creative writing, grammar, current news, and math. This set can be effortlessly expanded, simply by providing new categories for the models to generate. Our experiments show results aggregated across all categories, per-category rankings on a select number of topics is available in~\Cref{tab:topic_leaderboard}.

\begin{figure}[tbh]
  \includegraphics[width=0.98\columnwidth]{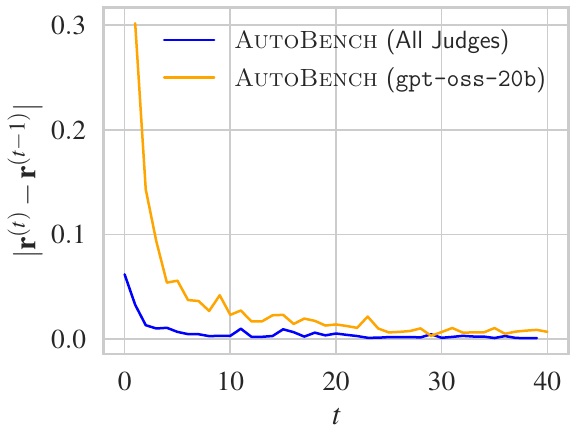}
  \caption{Convergence of weights over iterations, quantified by the L1 norm between successive evaluations.}
  \label{fig:l1_norm}
\end{figure}

\paragraph{Results and Discussion}
As shown in \Cref{tab:merged_corr_ablation}, \textsc{AutoBench} yields rankings that are well-aligned with recognized human-validated benchmarks. The multi-judge configuration achieves substantially higher correlation than the single-judge baseline, confirming that collective evaluation leads to more robust and human-consistent judgments. Strong alignment with GPQA demonstrates the framework's ability to generalize beyond knowledge-based tests to assess deeper reasoning.

The weighting dynamics also show rapid convergence, as illustrated in \Cref{fig:l1_norm}. The L1 norm between successive judging weights decreases quickly, indicating that the iterative process stabilizes as authority scores self-organize toward a consensus equilibrium. By contrast, the single-judge variant exhibits noisier convergence, underscoring the stabilizing effect of distributed judgment aggregation.

Comprehensive supplementary materials, including the aggregate cross-model judgment matrix (\Cref{fig:cross_evaluation_summary}), detailed per-topic results (\Cref{tab:topic_leaderboard}), and all prompts, are available in the appendix.

\section{Conclusions and Future Work} 

We introduced \textsc{AutoBench}, a fully automated framework for LLM evaluation through reciprocal peer assessment. Our experiments validate this paradigm, demonstrating that its consensus-driven rankings correlate strongly with established benchmarks such as MMLU-Pro and GPQA. This confirms that a collective of LLMs can produce a reliable measure of capability without human supervision or a fixed ground truth. Crucially, our ablation study highlights the value of the multi-judge design, which proves more robust than a single-judge baseline by mitigating individual model biases. 

By successfully automating the entire evaluation lifecycle, \textsc{AutoBench} provides a scalable, contamination-resistant, and adaptive alternative to static test sets, establishing a viable paradigm for the continuous assessment of evolving language models.

\section{Limitations} 
We acknowledge three primary limitations in the current design of \textsc{AutoBench}. First, its evaluation is confined to open-source models. The dynamics of peer evaluation and authority convergence could differ significantly when including leading proprietary LLMs, which possess distinct capability profiles and training methodologies. Second, as a fully autonomous system, \textsc{AutoBench} is susceptible to systemic bias. It risks creating an ``echo chamber'' where models reinforce shared weaknesses or converge on evaluation criteria that gradually diverge from human-defined notions of quality. Finally, our validation is indirect, relying on correlation with existing benchmarks. This approach, while informative, does not replace direct, large-scale human evaluation, which remains the gold standard for assessing nuanced aspects of quality like creativity, subtle reasoning, and factual accuracy.

\section*{Ethical Consideration} 
The autonomous nature of \textsc{AutoBench} introduces key ethical considerations. The reciprocal peer-evaluation mechanism risks amplifying shared societal biases present in the participating models, as they may reward one another for biased responses. Furthermore, the framework currently lacks explicit safety filters, posing a risk that models could generate and inadequately penalize harmful or toxic content. We therefore caution that the leaderboards reflect an emergent, inter-model consensus and should not be interpreted as a definitive or holistic measure of a model's quality, safety, or alignment. \textsc{AutoBench} is intended as a research tool to complement, not replace, other evaluation methods, particularly direct human oversight, and should not be the sole basis for high-stakes decisions like model deployment.
\section*{Acknowledgements}
This research is derived from the original open-source AutoBench project \cite{autobench2024project}, an eZecute S.R.L. initiative with the substantial financial and expertize support by Translated S.R.L.. The authors would like to acknowledge the foundational work of the original AutoBench team. This paper was focused on bringing strong scientific validation to the project's core methodology.

\bibliography{custom}

\begin{thebibliography}{21}
\providecommand{\natexlab}[1]{#1}

\bibitem[{{AutoBench}(2024)}]{autobench2024hf}
{AutoBench}. 2024.
\newblock Autobench 1.0: A collective-llm-as-a-judge benchmark system.
\newblock \url{https://huggingface.co/AutoBench}.
\newblock Accessed: 2025-10-21.

\bibitem[{Chan et~al.(2023)Chan, Chen, Su, Yu, Xue, Zhang, Fu, and
  Liu}]{chan2023chatevalbetterllmbasedevaluators}
Chi-Min Chan, Weize Chen, Yusheng Su, Jianxuan Yu, Wei Xue, Shanghang Zhang,
  Jie Fu, and Zhiyuan Liu. 2023.
\newblock \href {https://arxiv.org/abs/2308.07201} {Chateval: Towards better
  llm-based evaluators through multi-agent debate}.
\newblock \emph{Preprint}, arXiv:2308.07201.

\bibitem[{Chiang et~al.(2024)Chiang, Zheng, Sheng, Angelopoulos, Li, Li, Zhu,
  Zhang, Jordan, Gonzalez et~al.}]{chiang2024chatbot}
Wei-Lin Chiang, Lianmin Zheng, Ying Sheng, Anastasios~Nikolas Angelopoulos,
  Tianle Li, Dacheng Li, Banghua Zhu, Hao Zhang, Michael Jordan, Joseph~E
  Gonzalez, and 1 others. 2024.
\newblock Chatbot arena: An open platform for evaluating llms by human
  preference.
\newblock In \emph{Forty-first International Conference on Machine Learning}.

\bibitem[{Cobbe et~al.(2021)Cobbe, Kosaraju, Bavarian, Chen, Jun, Kaiser,
  Plappert, Tworek, Hilton, Nakano, Hesse, and Schulman}]{cobbe2021gsm8k}
Karl Cobbe, Vineet Kosaraju, Mohammad Bavarian, Mark Chen, Heewoo Jun, Lukasz
  Kaiser, Matthias Plappert, Jerry Tworek, Jacob Hilton, Reiichiro Nakano,
  Christopher Hesse, and John Schulman. 2021.
\newblock Training verifiers to solve math word problems.
\newblock \emph{arXiv preprint arXiv:2110.14168}.

\bibitem[{{eZecute S.R.L.}(2024)}]{autobench2024project}
{eZecute S.R.L.} 2024.
\newblock Autobench: Delivering transparency in llm benchmarking.
\newblock \url{https://autobench.org/}.
\newblock Accessed: 2025-10-21.

\bibitem[{Frick et~al.(2025)Frick, Chen, Tennyson, Li, Chiang, Angelopoulos,
  and Stoica}]{frick2025prompttoleaderboard}
Evan Frick, Connor Chen, Joseph Tennyson, Tianle Li, Wei-Lin Chiang,
  Anastasios~N. Angelopoulos, and Ion Stoica. 2025.
\newblock \href {https://arxiv.org/abs/2502.14855} {Prompt-to-leaderboard}.
\newblock \emph{Preprint}, arXiv:2502.14855.

\bibitem[{Hendrycks et~al.(2021{\natexlab{a}})Hendrycks, Burns, Basart, Zou,
  Mazeika, Song, and
  Steinhardt}]{hendrycks2021measuringmassivemultitasklanguage}
Dan Hendrycks, Collin Burns, Steven Basart, Andy Zou, Mantas Mazeika, Dawn
  Song, and Jacob Steinhardt. 2021{\natexlab{a}}.
\newblock \href {https://arxiv.org/abs/2009.03300} {Measuring massive multitask
  language understanding}.
\newblock \emph{Preprint}, arXiv:2009.03300.

\bibitem[{Hendrycks et~al.(2021{\natexlab{b}})Hendrycks, Burns, Kadavath,
  Arora, Basart, Tang, Song, and Steinhardt}]{math_dataset}
Dan Hendrycks, Collin Burns, Saurav Kadavath, Akul Arora, Steven Basart, Eric
  Tang, Dawn Song, and Jacob Steinhardt. 2021{\natexlab{b}}.
\newblock \href {https://arxiv.org/abs/2103.03874} {Measuring mathematical
  problem solving with the {MATH} dataset}.
\newblock \emph{CoRR}, abs/2103.03874.

\bibitem[{Holtzman et~al.(2019)Holtzman, Buys, Du, Forbes, and
  Choi}]{nucleus_sampling}
Ari Holtzman, Jan Buys, Li~Du, Maxwell Forbes, and Yejin Choi. 2019.
\newblock \href {https://api.semanticscholar.org/CorpusID:127986954} {The
  curious case of neural text degeneration}.
\newblock \emph{ArXiv}, abs/1904.09751.

\bibitem[{Kendall(1938)}]{kendall}
M.~G. Kendall. 1938.
\newblock \href {https://doi.org/10.1093/biomet/30.1-2.81} {A new measure of
  rank correlation}.
\newblock \emph{Biometrika}, 30(1-2):81--93.

\bibitem[{Kiela et~al.(2021)Kiela, Bartolo, Nie, Kaushik, Geiger, Wu, Vidgen,
  Prasad, Singh, Ringshia, Ma, Thrush, Riedel, Waseem, Stenetorp, Jia, Bansal,
  Potts, and Williams}]{kiela2021dynabenchrethinkingbenchmarkingnlp}
Douwe Kiela, Max Bartolo, Yixin Nie, Divyansh Kaushik, Atticus Geiger,
  Zhengxuan Wu, Bertie Vidgen, Grusha Prasad, Amanpreet Singh, Pratik Ringshia,
  Zhiyi Ma, Tristan Thrush, Sebastian Riedel, Zeerak Waseem, Pontus Stenetorp,
  Robin Jia, Mohit Bansal, Christopher Potts, and Adina Williams. 2021.
\newblock \href {https://arxiv.org/abs/2104.14337} {Dynabench: Rethinking
  benchmarking in nlp}.
\newblock \emph{Preprint}, arXiv:2104.14337.

\bibitem[{Kruger(2025{\natexlab{a}})}]{kruger2025autobenchrun3}
Peter Kruger. 2025{\natexlab{a}}.
\newblock Autobench third run: Revolutionizing llm evaluation with
  record-breaking scale, accuracy, and a new home at autobench.org.
\newblock \url{https://huggingface.co/blog/PeterKruger/autobench-run-3}.
\newblock Accessed: 2025-10-21.

\bibitem[{Kruger(2025{\natexlab{b}})}]{kruger2025autobench}
Peter Kruger. 2025{\natexlab{b}}.
\newblock Escape the benchmark trap: Autobench – the
  collective-llm-as-a-judge system for evaluating ai models (asi-ready!).
\newblock \url{https://huggingface.co/blog/PeterKruger/autobench}.
\newblock Accessed: 2025-10-21.

\bibitem[{Li and Murr(2024)}]{li2024humaneval}
Daniel Li and Lincoln Murr. 2024.
\newblock Humaneval on latest gpt models--2024.
\newblock \emph{arXiv preprint arXiv:2402.14852}.

\bibitem[{Li et~al.(2024)Li, Chiang, Frick, Dunlap, Wu, Zhu, Gonzalez, and
  Stoica}]{li2024crowdsourceddatahighqualitybenchmarks}
Tianle Li, Wei-Lin Chiang, Evan Frick, Lisa Dunlap, Tianhao Wu, Banghua Zhu,
  Joseph~E. Gonzalez, and Ion Stoica. 2024.
\newblock \href {https://arxiv.org/abs/2406.11939} {From crowdsourced data to
  high-quality benchmarks: Arena-hard and benchbuilder pipeline}.
\newblock \emph{Preprint}, arXiv:2406.11939.

\bibitem[{OpenAI(2023)}]{openai2023gpt4}
OpenAI. 2023.
\newblock \href {https://arxiv.org/abs/2303.08774} {Gpt-4 technical report}.
\newblock \emph{Preprint}, arXiv:2303.08774.

\bibitem[{Rein et~al.(2024)Rein, Hou, Stickland, Petty, Pang, Dirani, Michael,
  and Bowman}]{rein2024gpqa}
David Rein, Betty~Li Hou, Asa~Cooper Stickland, Jackson Petty, Richard~Yuanzhe
  Pang, Julien Dirani, Julian Michael, and Samuel~R Bowman. 2024.
\newblock Gpqa: A graduate-level google-proof q\&a benchmark.
\newblock In \emph{First Conference on Language Modeling}.

\bibitem[{Spearman(2015)}]{Spearman2015ThePA}
C.~Spearman. 2015.
\newblock \href {https://api.semanticscholar.org/CorpusID:14780428} {The proof
  and measurement of association between two things.}
\newblock \emph{International journal of epidemiology}, 39 5:1137--50.

\bibitem[{White et~al.(2025)White, Dooley, Roberts, Pal, Feuer, Jain,
  Shwartz-Ziv, Jain, Saifullah, Dey, Shubh-Agrawal, Sandha, Naidu, Hegde,
  LeCun, Goldstein, Neiswanger, and
  Goldblum}]{white2025livebenchchallengingcontaminationlimitedllm}
Colin White, Samuel Dooley, Manley Roberts, Arka Pal, Ben Feuer, Siddhartha
  Jain, Ravid Shwartz-Ziv, Neel Jain, Khalid Saifullah, Sreemanti Dey,
  Shubh-Agrawal, Sandeep~Singh Sandha, Siddartha Naidu, Chinmay Hegde, Yann
  LeCun, Tom Goldstein, Willie Neiswanger, and Micah Goldblum. 2025.
\newblock \href {https://arxiv.org/abs/2406.19314} {Livebench: A challenging,
  contamination-limited llm benchmark}.
\newblock \emph{Preprint}, arXiv:2406.19314.

\bibitem[{Zhao et~al.(2024)Zhao, Zhang, Chia, Xu, Zhao, and
  Bing}]{zhao2024autoarenaautomatingllmevaluations}
Ruochen Zhao, Wenxuan Zhang, Yew~Ken Chia, Weiwen Xu, Deli Zhao, and Lidong
  Bing. 2024.
\newblock \href {https://arxiv.org/abs/2405.20267} {Auto-arena: Automating llm
  evaluations with agent peer battles and committee discussions}.
\newblock \emph{Preprint}, arXiv:2405.20267.

\bibitem[{Zheng et~al.(2023)Zheng, Chiang, Sheng, Zhuang, Wu, Zhuang, Lin, Li,
  Li, Xing, Zhang, Gonzalez, and
  Stoica}]{zheng2023judgingllmasajudgemtbenchchatbot}
Lianmin Zheng, Wei-Lin Chiang, Ying Sheng, Siyuan Zhuang, Zhanghao Wu, Yonghao
  Zhuang, Zi~Lin, Zhuohan Li, Dacheng Li, Eric~P. Xing, Hao Zhang, Joseph~E.
  Gonzalez, and Ion Stoica. 2023.
\newblock \href {https://arxiv.org/abs/2306.05685} {Judging llm-as-a-judge with
  mt-bench and chatbot arena}.
\newblock \emph{Preprint}, arXiv:2306.05685.

\end{thebibliography}

\appendix

\section{Appendix}\label{sec:appendix}

\paragraph{Prompts and Sampling Mechanisms}\label{sec:appendix_prompts}
This section of the appendix details the core prompts and sampling mechanisms that drive the \textsc{AutoBench} framework. 
First, a \texttt{difficulty} level is sampled, with the generated token being \texttt{`a very difficult'}, \texttt{`a difficult'}, or \texttt{`a'} (standard difficulty) with probabilities of $0.6$, $0.3$, and $0.1$, respectively, to favor more challenging problems. A \texttt{topic} is then sampled uniformly from the set: $\{$\text{math}, \text{current news}, \text{creative writing}, \text{logic}, \text{grammar}, \text{coding}, \text{history}, \text{general culture}, \text{science}, \text{technology}$\}$. After sampling a \texttt{topic} and \texttt{difficulty}, the framework constructs a sequence of task-specific prompts. 

The task prompt (\Cref{tab:prompt_question_gen}) instructs a model to formulate a clear, self-contained task. Once generated, all models act as contestants, producing answers following the guidelines in \Cref{tab:prompt_answer_gen}. Subsequently, models evaluate both task and answer quality. The task evaluation prompt (\Cref{tab:prompt_question_rank}) scores clarity, relevance, and appropriateness for the specified difficulty, while the answer evaluation prompt (\Cref{tab:prompt_answer_rank}) applies a strict correctness gate and ranks accuracy, clarity, and depth. Together, these structured prompts ensure coherent task generation, consistent answer formulation, and reliable model assessment throughout the \textsc{AutoBench} pipeline.

\paragraph{Pseudocode}
To ensure clarity and facilitate the reproducibility of our framework, we formalize the complete iterative procedure of \textsc{AutoBench} in \Cref{alg:autobench}. The algorithm details each phase of an iteration, from the initial task generation and its quality assurance check against score thresholds, to the subsequent peer evaluation where all models judge all answers. This pseudocode provides a complete blueprint of the methodology presented in this paper.

\paragraph{Model Suite Details}
To provide context for our experiments, we present the full list of language models evaluated in this study in \Cref{tab:model-card}. The selection was curated to include a diverse suite of contemporary open-source models, varying significantly in parameter count, architecture, and developer. This diversity is intended to ensure a comprehensive and robust assessment of the \textsc{AutoBench} framework's ability to differentiate between a wide range of model capabilities.

\paragraph{Supplementary Results}
We provide some supplementary results to complement the analysis in the main paper. \Cref{fig:cross_evaluation_summary} offers a granular view of the inter-model evaluation dynamics through a heatmap of the aggregate judgment matrix. We present the full aggregate leaderboard in \Cref{tab:aggregate_leaderboard}, showing the final \textsc{AutoBench} scores alongside their corresponding performance on MMLU-Pro and GPQA. Finally, \Cref{tab:topic_leaderboard} offers a fine-grained breakdown of model performance across different task categories, highlighting specific areas of strength and weakness.

\paragraph{LLM Usage Disclosure}
In accordance with ethical guidelines, the authors disclose the use of a large language model (\texttt{Gemini 2.5 Pro}) during the preparation of this paper. The model's application was confined to an advisory role for improving the clarity of the manuscript. The conceptual framework, experimental results, and all scientific claims are the exclusive contributions of the authors, who retain full responsibility for the work.

\paragraph{Computational Details}
The experiments were conducted on a single machine equipped with four NVIDIA A6000 GPUs. All models were hosted locally and served using the vLLM inference engine for optimized throughput. A full run of the \textsc{AutoBench} framework, comprising $40$ iterations, required approximately $30$ hours of computation time. Due to our limited memory budget, models were loaded into and unloaded from GPU memory between evaluation steps, with this process accounting for a significant portion of the total runtime.

\begin{algorithm*}[tbh]
\caption{Autobench Evaluation Algorithm}
\label{alg:autobench}
\begin{algorithmic}[1]
\Require Models $\mathcal{M}=\{M_1,\dots,M_n\}$, iterations $T$, thresholds $\lambda_{\mathrm{mean}}, \lambda_{\mathrm{median}}$, max retries $k$.
\Ensure Final scores $\mathbf{c}^{(T)}$ and weights $\mathbf{w}^{(T+1)}$.

\State \textbf{Initialize:} $\mathbf{c}^{(0)} \gets \mathbf{0}$; \quad $\mathbf{w}^{(1)} \gets (1/n,\dots,1/n)$

\For{$t=1,\dots,T$}
    \State $retries \gets 0$; \quad $task\_accepted \gets \mathrm{false}$
    \Repeat
        \State $M_q \gets sample(\mathcal{M})$
        \State $q^{(t)} \gets generate(M_q)$
  \State $\mathbf{s} \gets \big( score(M_1, q^{(t)}), \dots, score(M_n, q^{(t)}) \big)$
        \State $\phi_1 \gets \, w\_mean(\mathbf{s},\mathbf{w}^{(t)}) \ge \lambda_{\mathrm{mean}}$
        \State $\phi_2 \gets \, w\_median(\mathbf{s},\mathbf{w}^{(t)}) \ge \lambda_{\mathrm{median}}$
        \If{$\phi_1 \land \phi_2$}
            \State $task\_accepted \gets \mathrm{true}$
        \Else
            \State $retries \gets retries + 1$
        \EndIf
    \Until{$task\_accepted$ \textbf{or} $retries \ge k$}

    \If{$task\_accepted$}
        \State $\forall i,j\in\{1,\dots,n\},\; J^{(t)}_{i,j} \gets \texttt{score}\big(M_i,\texttt{answer}(M_j,q^{(t)})\big)$
        \State $\mathbf{r}^{(t)} \gets (J^{(t)})^\top \mathbf{w}^{(t)}$
        \State $\mathbf{c}^{(t)} \gets \dfrac{(t-1)\mathbf{c}^{(t-1)} + \mathbf{r}^{(t)}}{t}$
        \State $\mathbf{w}^{(t+1)} \gets \mathbf{c}^{(t)} / \sum_{i=1}^n c_i^{(t)}$
    \Else
        \State $\mathbf{c}^{(t)} \gets \mathbf{c}^{(t-1)}$; \quad $\mathbf{w}^{(t+1)} \gets \mathbf{w}^{(t)}$
    \EndIf
\EndFor
\State \Return $\mathbf{c}^{(T)}, \mathbf{w}^{(T+1)}$
\end{algorithmic}
\end{algorithm*}

\begin{table*}
  \centering
  \caption{
    Overview of the models evaluated on the AutoBench benchmark.
  }
  \label{tab:model-card}
  \resizebox{\textwidth}{!}{%
  \begin{tabular}{l r c c c}
    \toprule
    \textbf{Model} & \textbf{Parameters} & \textbf{Reasoning} & \textbf{Release} &\textbf{Access/License}  \\
    \midrule
    \texttt{SmolLM2-1.7B-Instruct} & \SI{1.71}{B} & \textcolor{red}{\XSolidBrush} & $2025$ & \texttt{apache-2.0} \\
    \texttt{Llama-3.2-1B-Instruct} & \SI{1.24}{B} & \textcolor{red}{\XSolidBrush} & $2024$ & \texttt{llama3.2} \\
    \texttt{Llama-3.2-3B-Instruct} & \SI{3.21}{B} & \textcolor{red}{\XSolidBrush} & $2024$ & \texttt{llama3.2} \\
    \texttt{Qwen2.5-3B-Instruct} & \SI{3.09}{B} & \textcolor{red}{\XSolidBrush} & $2024$ & \texttt{qwen} \\
    \texttt{Qwen2.5-7B-Instruct} & \SI{8.29}{B} & \textcolor{red}{\XSolidBrush} & $2025$ & \texttt{apache-2.0} \\
    \texttt{Qwen2.5-14B-Instruct} & \SI{14.80}{B} & \textcolor{red}{\XSolidBrush} & $2024$ & \texttt{apache-2.0} \\
    \texttt{Meta-Llama-3-8B-Instruct} & \SI{8.03}{B} & \textcolor{red}{\XSolidBrush} & $2024$ & \texttt{llama3.1} \\
    \texttt{gemma-7b-it} & \SI{8.54}{B} & \textcolor{red}{\XSolidBrush} & $2024$ & gemma \\
    \texttt{Mistral-Small-3.1-24B-Instruct-2503} & \SI{24.0}{B} & \textcolor{red}{\XSolidBrush} & $2025$ & \texttt{apache-2.0} \\
    \texttt{gpt-oss-20b} & \SI{21.50}{B} & \textcolor{green}{\CheckmarkBold} & $2025$ & \texttt{apache-2.0} \\
    \texttt{gemma-3-1b-it} & \SI{1.00}{B} & \textcolor{red}{\XSolidBrush} & $2025$ & \texttt{gemma} \\
    \texttt{Phi-3-mini-4k-instruct} & \SI{3.82}{B} & \textcolor{red}{\XSolidBrush} & $2024$ & \texttt{MIT} \\
    \bottomrule
  \end{tabular}
  }
\end{table*}

\begin{figure*}[tbh]
  \includegraphics[width=0.98\textwidth]{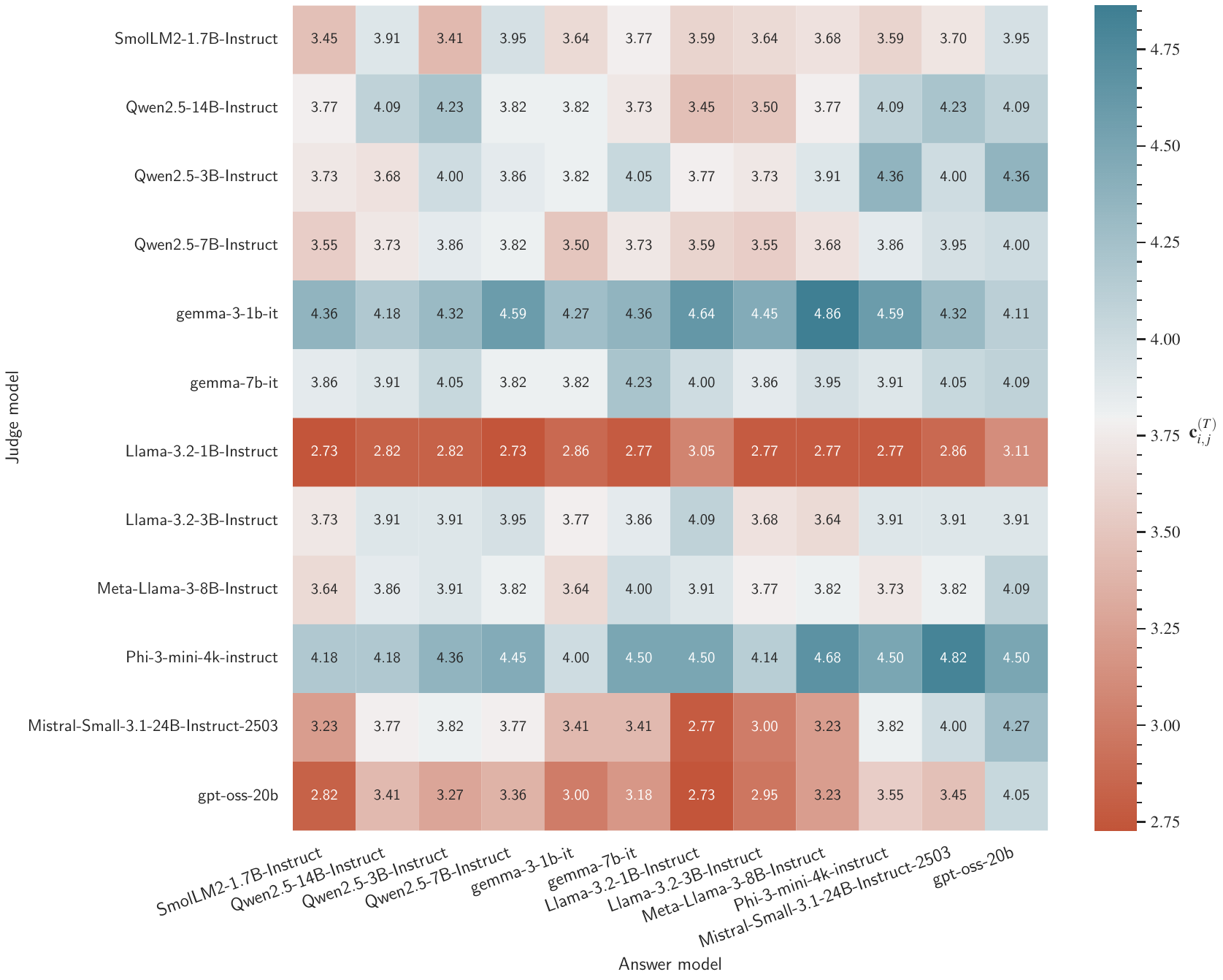}
  \caption{Heatmap of the aggregate judgment matrix, averaged over all $T$ iterations, each cell $(r, c)$ shows the mean score assigned by judge $r$ to the contestant model $c$. The main diagonal $(r = c)$ reveals potential self-preference, as it shows the score each model gives to itself.}
  \label{fig:cross_evaluation_summary}
\end{figure*}

\begin{table*}[tbh]
  \centering
  \caption{Aggregate model performance across benchmarks. We report the primary score for each benchmark.}
  \label{tab:aggregate_leaderboard}
  \resizebox{\textwidth}{!}{%
  \begin{tabular}{l c c c}
    \toprule
    \textbf{Model} & \textbf{AutoBench (Ours)} & \textbf{MMLU-Pro} &  \textbf{GPQA} \\
    \midrule
    \texttt{SmolLM2-1.7B-Instruct} & $3.80$ & $11.71\%$ & $3.91\%$ \\
    \texttt{Llama-3.2-1B-Instruct} & $3.58$ & $7.58\%$  & $3.36\%$ \\
    \texttt{Llama-3.2-3B-Instruct} & $3.73$ & $24.39\%$  & $3.80\%$ \\
    \texttt{Qwen2.5-3B-Instruct} & $3.77$ & $25.05\%$ & $3.02\%$ \\
    \texttt{Qwen2.5-7B-Instruct} & $3.80$ & $36.52\%$ & $5.48\%$ \\
    \texttt{Qwen2.5-14B-Instruct} & $3.91$ & $43.38\%$ & $9.62\%$ \\
    \texttt{Meta-Llama-3-8B-Instruct} & $3.85$ & $29.60\%$ & $1.23\%$ \\
    \texttt{gemma-7b-it} & $3.85$ & $7.72\%$  & $4.59\%$ \\
    \texttt{Mistral-Small-3.1-24B-Instruct-2503} & $3.95$ & $66.76\%$ &  $44.42\%$ \\
    \texttt{gpt-oss-20b} & $4.17$ & $73.14\%$ &  $71.50\%$ \\
    \texttt{gemma-3-1b-it} & $3.77$ & $14.70\%$ &  $19.20\%$ \\
    \texttt{Phi-3-mini-4k-instruct} & $3.91$ & $33.58\%$ &  $10.96\%$ \\
    \bottomrule
  \end{tabular}
  }
\end{table*}

\begin{table*}[tbh]
  \centering
  \caption{Detailed score overview of model performance on the different categories within the AutoBench benchmark. This breakdown highlights specific strengths and weaknesses of each model.}
  \label{tab:topic_leaderboard}
  \resizebox{\textwidth}{!}{%
  \begin{tabular}{l c c c c c c}
    \toprule
    \textbf{Model} & \textbf{General Culture} & \textbf{Grammar} & \textbf{History} & \textbf{Math} & \textbf{Science} & \textbf{Technology} \\
    \midrule
    \texttt{SmolLM2-1.7B-Instruct} & $4.06$ & $3.75$ & $3.87$ & $3.02$ & $3.52$ & $3.75$ \\
    \texttt{Llama-3.2-1B-Instruct} & $4.08$ & $3.57$ & $3.60$ & $3.79$ & $3.28$ & $3.70$ \\
    \texttt{Llama-3.2-3B-Instruct} & $4.06$ & $2.32$ & $3.84$ & $3.47$ & $3.12$ & $4.00$ \\
    \texttt{Qwen2.5-3B-Instruct} & $4.08$ & $3.74$ & $3.83$ & $3.80$ & $3.71$ & $3.83$ \\
    \texttt{Qwen2.5-7B-Instruct} & $4.13$ & $3.42$ & $3.91$ & $3.71$ & $3.70$ & $3.91$ \\
    \texttt{Qwen2.5-14B-Instruct} & $3.95$ & $3.75$ & $3.90$ & $3.44$ & $3.98$ & $3.87$ \\
    \texttt{Meta-Llama-3-8B-Instruct} & $4.02$ & $3.83$ & $3.91$ & $3.58$ & $3.66$ & $3.70$ \\
    \texttt{gemma-7b-it} & $4.11$ & $3.67$ & $3.81$ & $4.03$ & $3.12$ & $4.00$ \\
    \texttt{Mistral-Small-3.1-24B-Instruct-2503} & $4.00$ & $4.00$ & $3.96$ & $3.76$ & $4.06$ & $3.92$ \\
    \texttt{gpt-oss-20b} & $4.45$ & $3.93$ & $4.08$ & $3.70$ & $4.14$ & $4.29$ \\
    \texttt{gemma-3-1b-it} & $4.08$ & $1.82$ & $3.86$ & $3.75$ & $3.11$ & $3.79$ \\
    \texttt{Phi-3-mini-4k-instruct} & $4.11$ & $2.32$ & $3.94$ & $3.96$ & $3.83$ & $4.09$ \\
    \bottomrule
  \end{tabular}
  }
\end{table*}

\begin{table*}[tbh]
\centering
\caption{Prompt for Task Generation.}
\label{tab:prompt_question_gen}
\begin{tabular}{p{0.9\linewidth}}
	\toprule
	\textbf{Task Generation} \\ \midrule
\begin{minipage}[t]{\linewidth}
\begin{verbatim}
Generate {difficulty} question about {topic}. Keep the question clear, 
self-contained, and unambiguous.

Requires analysis and multi-step reasoning; expect intermediate 
difficulty.

Pose a math problem (word problem or symbolic) that requires 
calculation or reasoning. Include units if relevant.

Return exactly the final question text only - do NOT include 
chain-of-thought, commentary, steps, or explanations.

Do not surround the question with quotes or extraneous punctuation. 
Keep it concise and well-formed.

If you are a thinking/reasoning model, ALWAYS ensure the final 
question appears in the standard 'content' field only; do NOT 
place it in 'reasoning_content', do NOT leave the 'content' 
field empty once you finish reasoning.

\end{verbatim}
\end{minipage}
\\
\bottomrule
\end{tabular}
\end{table*}

\begin{table*}[tbh]
\centering
\caption{Prompt for Answer Generation.}
\label{tab:prompt_answer_gen}
\begin{tabular}{p{0.9\linewidth}}
	\toprule
	\textbf{Answer Generation} \\
\midrule
\begin{minipage}[t]{\linewidth}
\begin{verbatim}
Answer the question clearly, correctly, and concisely. Avoid 
hallucinations and speculative claims; include only facts you 
are confident about. Do NOT reveal chain-of-thought, internal 
reasoning steps, or chain-of-thought tags; output only the final 
answer. If you are a thinking/reasoning model, ALWAYS place the 
final answer in the standard 'content' field only; do NOT place 
it in 'reasoning_content', do NOT leave the 'content' field 
empty once you finish reasoning.

Question: [Question text is inserted here]

Answer:

\end{verbatim}
\end{minipage}
\\
\bottomrule
\end{tabular}
\end{table*}

\begin{table*}[tbh]
\centering
\caption{Prompt for Ranking a Task's Quality.}
\label{tab:prompt_question_rank}
\begin{tabular}{p{0.9\linewidth}}
	\toprule
	\textbf{Task Rating} \\
\midrule
\begin{minipage}[t]{\linewidth}
\begin{verbatim}
Evaluate the quality of the question on an integer scale 1–5. 
Consider clarity, appropriateness for the stated difficulty, 
relevance to the topic, grammar/formulation, and engagement.

Return ONLY the final score as XML exactly like this: <rank>n</rank> 
with no extra text. If you are a thinking/reasoning model, place 
the final <rank> tag in the 'content' field only; do not place 
it in 'reasoning_content'.

Question: [Question text is inserted here]
Difficulty: [Difficulty label, e.g., 'difficult']
Topic: [Topic label, e.g., 'math']

Return:

\end{verbatim}
\end{minipage}
\\
\bottomrule
\end{tabular}
\end{table*}

\begin{table*}[tbh]
\centering
\caption{Prompt for Ranking an Answer's Quality.}
\label{tab:prompt_answer_rank}
\begin{tabular}{p{0.9\linewidth}}
	\toprule
	\textbf{Answer Rating} \\
\midrule
\begin{minipage}[t]{\linewidth}
\begin{verbatim}
You are an expert evaluator. Rank the answer on an integer scale 
1–5 using the criteria below. **Strict correctness gate:** if the 
answer contains any factual error, clear hallucination, or an 
answer that is broken/irrelevant for the question, the maximum 
score allowed is 2. After applying the gate, judge on 
correctness, clarity, relevance, structure/grammar, and depth.

Scale (post-gate):
1: Poor/off-topic or contains substantial errors.
2: Partially correct but contains important mistakes or omissions.
3: Correct, clear, and relevant; adequate but not deep.
4: Correct with good structure, depth, and precise coverage.
5: Correct and exceptional — excellent clarity, original insight, 
   or thoroughness.

IMPORTANT: Return ONLY the rank wrapped in XML exactly like this: 
<rank>n</rank> where n is an integer 1-5. No additional text, 
no punctuation, no explanation. If you are a thinking/reasoning 
model, place the final <rank> tag in the 'content' field only; 
do not place it in 'reasoning_content'.

Question: [Question text is inserted here]
Answer: [Answer text is inserted here]

Return:

\end{verbatim}
\end{minipage}
\\
\bottomrule
\end{tabular}
\end{table*}

\end{document}